\documentclass[conference]{IEEEtran}
\IEEEoverridecommandlockouts

\usepackage{cite}
\usepackage{amsmath,amssymb,amsfonts}
\usepackage{algorithmic}
\usepackage{graphicx}
\usepackage{textcomp}
\usepackage{xcolor}
\usepackage{colortbl}
\usepackage{multirow}
\def\BibTeX{{\rm B\kern-.05em{\sc i\kern-.025em b}\kern-.08em
    T\kern-.1667em\lower.7ex\hbox{E}\kern-.125emX}}

\usepackage{amssymb}
\usepackage{pifont}
\usepackage{orcidlink}
\usepackage{enumitem}
\setlist{nosep}
\usepackage{multirow}
\usepackage{booktabs}
\usepackage{subcaption}
\usepackage{hyperref}
\usepackage{balance}
\usepackage[normalem]{ulem}
\useunder{\uline}{\ul}{}
\hypersetup{
    colorlinks=true,
    linkcolor=blue,
    filecolor=magenta,      
    urlcolor=cyan,
}
\definecolor{lightgray}{gray}{0.9}

\usepackage{xspace}
\makeatletter
\DeclareRobustCommand\onedot{\futurelet\@let@token\@onedot}
\def\@onedot{\ifx\@let@token.\else.\null\fi\xspace}

\begin{document}

\title{Evaluation of Image Matching for Art Skills Assessment}


\author{
\IEEEauthorblockN{
Asaad Alghamdi\textsuperscript{\rm 1}, Michael Poor\textsuperscript{\rm 1}, Trung-Nghia Le\textsuperscript{2,3}\orcidlink{0000-0002-7363-2610}, Tam V. Nguyen\textsuperscript{\rm 1}\orcidlink{0000-0003-0236-7992} \\
}

\IEEEauthorblockA{%
\textsuperscript{\rm 1}
University of Dayton, Ohio, United States\\%
\textsuperscript{\rm 2}
University of Science, VNU-HCM, Ho Chi Minh City, Vietnam\\%
\textsuperscript{\rm 3}
Vietnam National University, Ho Chi Minh City, Vietnam\\%
\textit{alghamdia41@udayton.edu, mpoor1@udayton.edu, ltnghia@fit.hcmus.edu.vn, tamnguyen@udayton.edu}
\ }
}

\maketitle

\begin{abstract}

While some individuals possess a natural talent for drawing, mastering this skill requires dedicated training and practice. Determining one's skill in the art of drawing requires proper comprehensive assessment. In this paper, we propose a method to measure drawing skill by by matching the hand-drawn image with the original template. Existing techniques often involve complex processes. However, advancements in computer vision allow us to train computers to perform these comparisons at a human-like level, thereby resolving the tedious and overwhelming traditional process. Using computer vision applications, determining image similarity involves identifying the level of similarities in an image with a reference image. We have implemented and analyzed the SIFT feature and Siamese network to measure image similarity. Our results indicate that it is feasible to assess art skill levels. Through feature analysis, we found that SIFT-based key point matching provides a more effective means of detecting drawing skills.

\end{abstract}


\section{Introduction}

As technology continues to advance, we are entering an era where computers can perform tasks that traditionally required human expertise. Historically, researchers have noted that computers offer significant advantages over humans, such as superior memory capabilities, the ability to store vast amounts of information, and the capacity to operate continuously without the need for rest, enabling them to perform calculations and analysis tirelessly around the clock.

This paper aims to develop a framework that measures artistic skills by quantifying the accuracy of a user's sketch against a template photo. To this end, feature-matching techniques are incorporated to be evaluated to produce the most accurate results possible. The proposed computational framework utilizes feature matching to analyze artistic abilities, implemented through OpenCV \cite{lukac2018color}, a famous computer vision library. Particularly,  we evaluate how a computer vision-based model assesses art skills by comparing user-drawn sketches with actual template images. This comparison involves analyzing various elements of the images and producing a value that reflects the visual similarity between the sketch and the template. Our computational framework addresses the art assessment problem by comparing the similarities between a given template image and a corresponding drawn sketch. This is achieved through descriptor similarity, obtained by correlating small patches within the neighborhood of the images. This method eliminates the need for modality transformation, significantly reducing the intramodality gap. We utilize a database containing sketch-photo pairs and the actual images to validate our outcomes.

To verify the accuracy of our framework, we compare the results of SIFT \cite{lowe1999object, lowe2004distinctive} and Siamese networks \cite{melekhov2016siamese} with human assessments, which involves seeking expert opinions on the quality of the sketches. Our proposed framework can scan drawings and produce a numerical value indicating the aesthetic level of the artwork. It demonstrates the feasibility of using computer vision techniques to assess artistic skills, providing a valuable tool for artists and educators to evaluate and improve drawing proficiency.

\begin{figure}[t!]
    \centering
    \includegraphics[width=\linewidth]{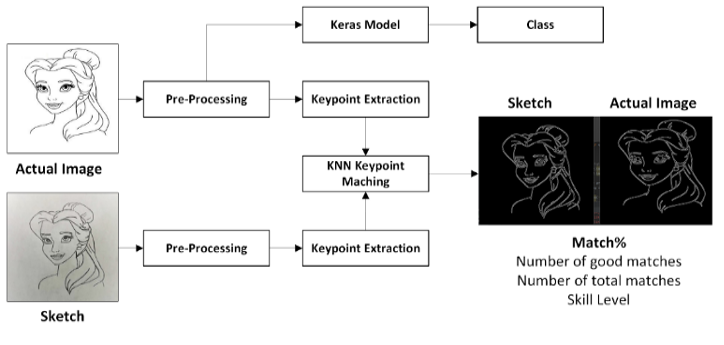}
    \caption{Visualization of matching between a template and a sketch by using SIFT keypoints.}
    \label{fig:matching}
\end{figure}

The remainder of this paper is organized as follows. Section \ref{sec:related_work} briefly summarizes the related work in literature. Next, Section \ref{sec:proposed_method} introduces the methodology for art skills assessment. Section \ref{sec:experiments} presents the experimental results along with discussion. Finally, Section \ref{sec:conclusion} concludes the paper and paves the way for future work.

\begin{figure*}[t!]
    \centering
    \includegraphics[width=0.75\linewidth]{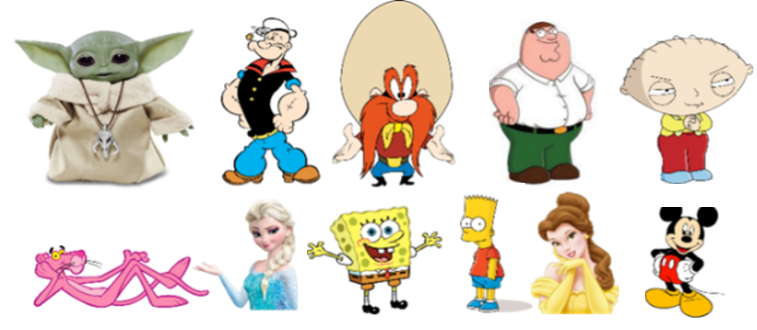}
    \caption{The collage of some images used in this research work.}
    \label{fig:data}
\end{figure*}

\section{Related Work}
\label{sec:related_work}

As aforementioned, our aim is to create a program to assess a hand-drawn object and evaluate the artistic talent of a user. In fact, Chie \cite{chie} mentioned that the technology of computer vision assessment may improve the art skills of humans if adopted. Using computer vision techniques to measure art skills is a field that was not well explored before. Valueva et al. \cite{valueva2020application} demonstrated various ways to assess the quality of a hand-drawn image using a computerized program. In their work, using a pre-determined scale of the results, sketches can be matched to a respective image to evaluate the skill level. Sethi \cite{sethi} discussed on how computers can be used to understand art. Using computer vision eliminates the need to rely on subjective perception and instead uses optimization techniques to discover novel insights into artwork. The color and varying features across different painting parts strongly indicate the artist's style. As discussed by Sethi \cite{sethi}, when an image is scanned, it was broken down into individual pixels with numeric values, such as how much blue, red, and green are in each section. The values obtained by the system are graphically represented to give us another view of the artwork.

In this work, we focus on locating keypoints and extracting corresponding descriptors of the pre-determined template image and the sketch. Bruno et al. \cite{bruno2013object} investigated two descriptors, namely, SIFT and ORB methods. Based on the performance evaluation, SIFT is considered rotation invariant and more noise-resistant. On the other hand, Karami et al. \cite{krizhevsky2012imagenet} conducted an extensive study of image matching by using SIFT, BRIEF, ORB, and SURF. They found that ORB is the fastest algorithm while SIFT performs the best in the most scenarios. Therefore, we also consider using SIFT in this work. 

In addition to the aforementioned handcrafted features, Convolutional Neural Networks (CNNs)  \cite{krizhevsky2012imagenet, redmon2016you,nguyen2019detection, long2015fully,nguyen2016segmentation,nguyen2019saliency} are widely used for image recognition, detection, segmentation. For example, VGG16, a popular CNN model, has trainable layers that make it possible to fit accurately in more complex functions. Melekhov et al. \cite{melekhov2016siamese} used VGG16 to train the model for evaluating the artistic skills of a sketch based on the original image as the point of reference. They suggest that VGG16 architecture is suitable for accurately performing complex functions. CNNs are also used in image matching. Lin et al. \cite{lin2015learning} studied how to use deep networks to geolocative images by matching images without using ground-level as a reference image. Hence, we plan to investigate both SIFT and CNN-based matching for this work.

\section{Methodology}
\label{sec:proposed_method}

Methods that are used to achieve research aims are outlined in this section. The main objective of the research is to develop a hand-drawing measuring model based on image matching. Specifically,we calculate the similarity between two images using SIFT as the feature extraction method. The extracted features are then fed into the Siamese Network based on VGG16 that uses Cosine and Euclidean similarity to classify further which class the images belong to. Our study attempts to explore which methods are suitable for certain art skill levels.

\subsection{SIFT Keypoint Matching}

Lowe \cite{lowe2004distinctive, lowe1999object} proposed SIFT keypoints to assist image rotation, affine transformations, changes in viewpoints, and intensity in matching features. The actual process involves four steps: estimating a scale-space extremum using Difference of Gaussians (DoG), localization of keypoints by eliminating low contrast points, orientation, and assignment of a keypoint based on local image gradient. Finally, a descriptor generator is used to compute the local image descriptor based on the magnitude and gradient of the image for each keypoint \cite{lowe2004distinctive}. 

We introduce a multi-stage matching framework based on SIFT. The framework involves many stages categorized into three major sections: detection of classes, art skill level detection, and classification of art skills. The flowchart of using SIFT keypoint for image matching is illustrated in Fig. \ref{fig:matching}.

\subsection{Template Dataset}

\begin{figure*}[t!]
    \centering
    \includegraphics[width=0.75\linewidth]{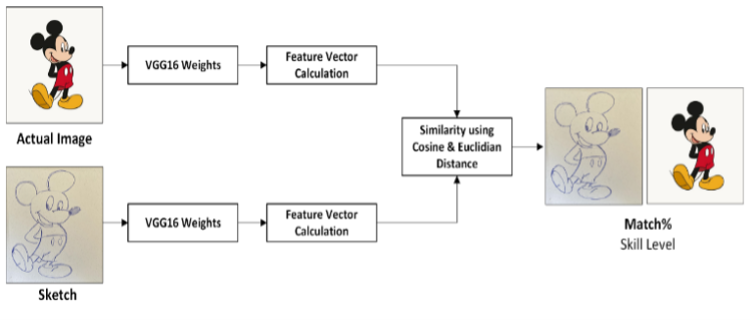}
    \caption{Siamese based image matching.}
    \label{fig:siamese}
\end{figure*}

In this work, 10 classes of different cartoon characters are being collected. The characters were selected by arts professors as those characters are familiar to the participants and do not require much time to draw. A collage of some images that are used in this research is demonstrated in Fig. \ref{fig:data}.

Data collection is the first step to train a model. We collect many sketches for each image template from a pool of 120 participants. We train the framework (also shown in Fig. \ref{fig:matching}) on Google Collab platform. When training is started initially, we have more data loss as weights are inaccurate. However, as training cycles increase, data loss decreases, and accuracy increases. The class detection model is loaded into the program, and a prediction is made. The input image to the model is our original image, which is also used to compare with the artwork.

\subsubsection{Art Skill Levels} 

For skill level detection, SIFT features and Flann-based K-neighbor search are utilized for art skill level identification. The following steps are used to identify skills level:

\textbf{Pre-Processing. } Before matching both images, there are pre-processed steps for finding keypoints. The following steps are included in pre-processing the image:

\begin{table}[t!]
\centering
\caption{Rating skills percentages for SIFT enhanced method.}
\label{tab:sift_matching}
\begin{tabular}{l|c}
\multicolumn{1}{c|}{\textbf{Skill levels}} & \textbf{Rating scales based on matching percentage} \\
\hline
Beginner & 0-65 \\
Intermediate & 65-80 \\
Advanced & 80-90 \\
Professional & 90-100
\end{tabular}
\end{table}

\begin{enumerate}
    \item Resizing images: the image is resized to be of size 500 × 500.
    \item Grey scaling: the image is converted from RGB to greyscale to reduce the color channels from 3 to 1 and minimize the processing.
    \item Gaussian Blur: The Gaussian Blur filter is applied to the image to blur the image to smoothen the edge detection process, dilation, and SIFT calculation.
    \item The auto canny function is calculated using the Sigma value of an image: The canny edge detection is applied to the image to convert the image into the binary image with edges in the white color.
    \item OpenCV trackbars to change image detail level in real-time in case of auto canny does not provide accurate results.
    \item In black and white images, broken lines or gaps are filled using the Erode Function.
    \item Morphological Operations, such as erode and dilate, are used where the value could be changed in real-time to suppress noise.
    \item Finally, an image is returned and passed to the SIFT matching flowchart.
\end{enumerate}

\textbf{Extracting Keypoints. } SIFT 2D features are used to detect and compute the key points of images. In particular, the nearby pixels' alignments and magnitudes are used to create a unique fingerprint for this keypoint called a 'descriptor.’

\textbf{Matching Keypoints. } By locating their nearest neighbors, keypoints between two images are matched. The second-closest match may be extremely close to the first in certain circumstances. This might occur due to noise or other causes. Nonetheless, the closest distance to the second-closest distance percentage, which is 98\%, is used; if it is more than a 98\% match, it is discarded. Keypoints of both images are matched using a Fast Library for Approximate Nearest Neighbors (Flann)-based K-neighbor search \cite{lukac2018color}. All matches are also shown in the plot to infer the results to help to judge the result since some noises in the image could be seen as a match.

\subsubsection{Art Skills Assessment }

This step takes two images and calculates the keypoints and descriptors for both images. The descriptors are then used for matching two images. Based on the number of matches, a score is calculated, showing how similar or different the images are from each other. Based on the calculated score, the art skill level is determined and printed on the screen along with the concatenated image with similar points. Based on match percentages, art skill levels are calculated following these roles, as shown in Table \ref{tab:sift_matching}.

\subsection{Siamese Network-based approach}

The Siamese neural network comprises two conjoined twin networks that receive different inputs but are connected by an energy function at the top \cite{long2015fully}. Bromley et al. \cite{bromley1993signature} introduced Siamese network technology in the early 1990s. The technology was meant to assist in signature verification through image matching. This technique has two main characteristics: 

\begin{itemize}
    \item Forecasts are guaranteed to be consistent. Since each network computes the same function, weight tying ensures their networks cannot translate two highly similar pictures to very different places in feature space.
    \item The network is symmetric. When showing two different pictures to the twin networks, the top conjoining layer calculates the same metric as if we showed the same two photos to the opposite twins.  
\end{itemize}

One of these networks receives each picture in the image pair. Similarity measures can be employed where a twin network would be useful, such as identifying handwritten checks, automated recognition of faces in camera pictures, and matching searches with indexed texts \cite{michelucci2019advanced}. Face recognition is likely the most well-known application of Siamese networks, in which known photos of individuals are efficiently matched to an image from a turnstile or similar.

The pre-trained VGG16 model uses the Cosine similarity to calculate vectors for both input and output. When a Siamese network is used to compare images, images to be compared are passed through subnetworks that share the weight. As a result, images that are categorized as the same class have identical 4096-dimensional representations. The output feature vector generated from each subnetwork is combined through subtraction, and the outcome is passed through an operation fully connected with a single output. A sigmoid operation converts these values to a probability ranging from 0 to 1, indicating whether images are similar or dissimilar.

\begin{table}[t!]
\centering
\caption{Rating skills percentages based on matching percentages.}
\label{tab:matching}
\begin{tabular}{l|c|c}
\multicolumn{1}{c|}{\textbf{Skill levels}} & \textbf{Cosine similarity \%} & \textbf{Euclidian distance} \\
\hline
Beginner & 0-70 & 0-65\\
Intermediate & 70-75 & 65-73 \\
Advanced & 75-83 & 73-79\\
Professional & 83-100 & 79-100
\end{tabular}
\end{table}

\textbf{VGG16-based model using Cosine similarity and Euclidean distance}

The proposed work finds the similarity between two images by employing transfer learning. Fig. \ref{fig:siamese} shows the framework of our second Siamese (VGG16) model. Our proposed method calculates the similarity percentage between two images using Cosine similarity and Euclidean distance.

\begin{equation}
    similarity(A,B)=\frac{A \cdot B}{\left \| A \right \| \times \left \| B \right \|},
\end{equation}
\begin{equation}
    d(p,q)=d(q,p)=\sqrt{\sum_{i=1}^{n}(q_i-p_i)^2}.
\end{equation}
 
Euclidean Distance uses the concept of transfer learning to extract the features from an image and calculate the distance instead of utilizing a histogram to extract features. Euclidean Distance involves the application of the pre-trained model to an existing problem by making minor adjustments to fit the required problems. The problem statement has been implemented using the Keras with TensorFlow. 

We also adopt CNNs to extract the features from the images by convolving the different filters of a specific size over the whole image. These filters extract the features, i.e., vertical and horizontal edges of objects from the images. After the feature extraction, the similarity between two images is calculated by finding mathematical measures, particularly Cosine similarity or Euclidean distance \cite{karami2017image}. 

In particular, we use the pre-trained VGG16 model imported from the Keras API by defining the input shape and importing the weights of the models that are trained on the ImageNet dataset. We read all images and extract their features. Then, the similarity calculation methods are called to compute the distances between image features. Finally, the percentage of the similarity is returned by the functions. We use the rating values in Table \ref{tab:matching} to determine the class label for each sketch.

\begin{table}[t!]
\centering
\caption{Analyzing results of sketches at different drawing levels.}
\label{tab:sketches}
\begin{tabular}{l|ccc}
 & \textbf{SIFT} & \begin{tabular}[c]{@{}c@{}}\textbf{VGG16}\\ \textbf{(Cosine)}\end{tabular} & \begin{tabular}[c]{@{}c@{}}\textbf{VGG16}\\ \textbf{(Euclidean)}\end{tabular} \\
 \hline
 & \multicolumn{3}{c}{\textbf{Beginner Level}} \\
 \hline
\textbf{Mean} & 55.89 & 65.42 & 61.37 \\
\textbf{Std} & 18.82 & 9.95 & 11.68 \\
\hline
 & \multicolumn{3}{c}{\textbf{Intermediate Level}} \\
 \hline
\textbf{Mean} & 77.97 & 78.57 & 73.45 \\
\textbf{Std} & 6.83 & 4.62 & 4.13 \\
\hline
 & \multicolumn{3}{c}{\textbf{Advanced Level}} \\
 \hline
\textbf{Mean} & 83.04 & 77.55 & 75.39 \\
\textbf{Std} & 4.48 & 3.62 & 3.34 \\
\hline
 & \multicolumn{3}{c}{\textbf{Professional Level}} \\
 \hline
\textbf{Mean} & 89.03 & 86.60 & 84.50 \\
\textbf{Std} & 4.92 & 5.47 & 4.56
\end{tabular}
\end{table}

\begin{figure*}[t!]
    \centering
    \includegraphics[width=\textwidth]{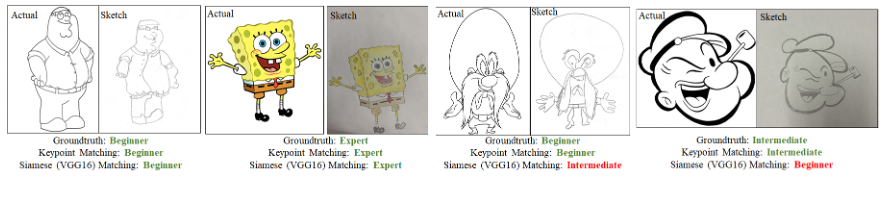}
    \caption{Examples of art skills assessments from keypoint matching and Siamese matching methods. Green and red indicate the correct and incorrect predictions, respectively.}
    \label{fig:visualization}
\end{figure*}

\subsection{Data Collection}

120 participants with differing drawing levels completed 189 sketches. Sketches that contain noise, distortion, and sharpness are removed manually to enhance accuracy during the process to ensure quality data. 

\subsection{Art Skill Levels}

Making the same skill level scales for both methods that are illustrated previously is not an effective approach as these models calculate similarity differently. To achieve an accurate approach to scale drawing skills, a dataset of 83 images was used in this step, and it was divided into four categories based on visuals, namely, beginner (25 images), intermediate (16), advanced (20), and professional (22). Furthermore, almost the same number of images from each class were present to have a balanced dataset. Every subclass contained each category of sketch class. All images were evaluated individually as there is no way to automate the system since every image's sketch class is changed. Additionally, we must change the source and target sketch every time. Each percentage value was recorded with respect to the similarity method, and results for each category were separated. Following this, statistical methods are leveraged to interpret our results from the experiment. Mean and standard deviation are two parameters that can assist us in choosing value ranges for skill level. Mean is an average value that tells us where our model values stand. This gives us a technique to obtain statistically designed accurate skill level scales. Table \ref{tab:sketches} shows the values for standard deviation and the average of each category, beginner, intermediate, advanced, and professional level for both methods (SIFT and Siamese - VGG16).

From the results in the previous tables, it could be observed that ranges are defined distinctively for both methods (SIFT and VGG16) except for the intermediate and advanced levels of Cosine similarity. Based on the mentioned method, Table \ref{tab:comparision} demonstrates the ranges for the art skill level scales we suggest for both enhanced methods.

\section{Results and Discussions}
\label{sec:experiments}

\begin{table}[t!]
\centering
\caption{Rating skills percentages for SIFT and Siamese (VGG16) matching models.}
\label{tab:comparision}
\begin{tabular}{l|ccc}
\multirow{2}{*}{\textbf{Skill Levels}} & \multicolumn{3}{c}{\textbf{Rating scales based on matching percentage \%}} \\
\cline{2-4}
 & \textbf{SIFT} & \textbf{Cosine similarity} & \textbf{Euclidian} \\
 \hline
Beginner & 0-65 & 0-70 & 0-65 \\
Intermediate & 65-80 & 70-75 & 65-73 \\
Advanced & 80-90 & 75-83 & 73-79 \\
Professional & 90-100 & 83-100 & 79-100
\end{tabular}
\end{table}

The experimental results were synchronous with the expected results. Paying attention to an image dataset, which consists of images representing a typical real-world scene, is important. In some cases, the image size must be modified to promote the efficiency and rate of algorithm performance. The system takes two images as input, then compares both images with multiple methods and gives a judgment based on similarity percentage, providing us with the skill level. Another function of the program is to detect the class of each art, out of the total ten classes only selected for the system. The system works on already captured pictures and a set of original pictures. Human measuring was used to categorize the algorithms' results as beginner, intermediate, advanced, or professional.

\subsection{Failure Modes and Actions on Failure}

We summarize the failure cases during the assessment. Firstly, exploring the adaptability of Canny edge detection parameters in real-time could enhance the system's ability to capture and process edges accurately. However, the challenge lies in the inability to verify the auto Canny parameters, especially when working with black and white processed images, which could potentially affect the reliability of edge detection. Meanwhile, color is used as the input in Siamese-VGG16. In SIFT keypoint matching, the nearest neighbor matching is used for only good keypoints. It is worth noting that unwanted keypoints may affect results. We can use OpenCV morphological operations to reduce noise in real time. Last but not least, double-lined sketches produce noise while using black-and-white processed images.

\subsection{Feature Detection Results}

During the implementation of the features, it was found that SIFT-based matching can detect more features (Keypoints Detection) compared to Siamese with VGG16 (Feature Vectors). It is not a surprise that SIFT detected many features in the image as it is proven to be a robust detector. However, it is worth noting that the SIFT-based matching detects features in areas that do not have sufficient information in most images. In this scenario, parts of the image considered good features are edges, highly contrasting areas, and corners. An exception to note, SIFT does not detect excellent features in some cases. However, it has more advantages, such as it is more accurate than any other descriptors, unaffected by rotation, scaling, or image brightness, and works with any image irrespective of color, unlike Siamese with VGG16, where color significantly impacts results and is less crucial. Moreover, SIFT-based matching provides correct skill assessment that closely aligns with human ratings.

\begin{figure}[t!]
    \centering
    \includegraphics[width=\linewidth]{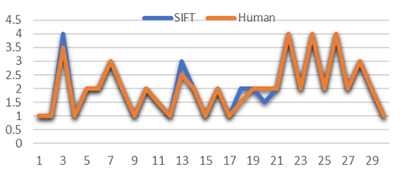}
    \caption{The comparison results between human and SIFT-based assessment.}
    \label{fig:result}
\end{figure}

The art assessment system outperforms other systems available in the market. Random 30 results are plotted, showing that SIFT is closest to human results, as shown in Fig. \ref{fig:result}. Fig. \ref{fig:visualization} shows examples of art skills assessment from both keypoint and Siamese matching methods. Indeed, SIFT detected above 95\% of keypoints related to each sketch and 70\% of good matches with their actual images. Analyzing the number of robust keypoints to survive further filtering is necessary. Furthermore, since these keypoints are scale- and orientation-invariant, they could provide results with more precision than alternative approaches. For implementation, a collection of pre-existing photos is kept in local storage and utilized to get keypoints. Since the sketch is always based on a picture, this set of photos may be found locally to complete the model. To compute the similarity, the model is fed with the original sketch, which is perfect, and the sketch we want to calculate similarity. Keypoints of both images are calculated and further processed to get similarities. Because they are unaffected by image size or orientation, SIFT features offer a substantial advantage over edge techniques. An overall accuracy of 80\%-98\% was attained using SIFT. This equated to a presumption that the samples' descending portions could be accurately categorized using this method. SIFT strategy was that SIFT is arguably the most preferred algorithm used to match under different scales, lighting, and rotation. However, compared to VGG16, SIFT is relatively slow. Even though it takes longer to examine all the findings, the SIFT approach is more accurate. KNN is utilized to identify good matches and eliminate mismatches. Moreover, both images can have different backgrounds and some noise, which may decrease the accuracy since an extra little line can have keypoints that may affect the results.

\section{Conclusion}
\label{sec:conclusion}

The overarching goal of this paper is to measure art skills automatically. The proposed work implements the calculation of image similarity between two images using SIFT as the feature extraction method and using the Siamese Network based on the pre-trained VGG16 model that uses Cosine and Euclidean similarity for feature vectors. In summary, from the results obtained from the study, it can be concluded that we can teach AI to assess human drawing skills. 

However, the main challenge is that the system cannot identify outliers and useful data. For future work, it is essential to prioritize refining outlier detection mechanisms and pre-processing techniques to bolster the accuracy and reliability of the automated art skill assessment system. Advanced outlier detection algorithms must be explored to filter noise and irrelevant data effectively, while further research into pre-processing methods such as noise reduction and image enhancement could optimize input data for improved performance across diverse drawing styles. Additionally, integrating additional feature extraction methods like SURF or ORB and exploring hybrid approaches combining traditional algorithms with deep learning architectures could offer enhanced insights.
\balance

\section*{Acknowledgement}

This research is supported by National Science Foundation (NSF) under Grant Number 2025234. Dr. Trung-Nghia Le is funded by Vietnam National Foundation for Science and Technology Development (NAFOSTED) under Grant Number 102.05-2023.31.

\bibliographystyle{IEEEtran}
\bibliography{mapr}

@book{lukac2018color,
  title={Color image processing: methods and applications},
  author={Lukac, Rastislav and Plataniotis, Konstantinos N},
  year={2018},
  publisher={CRC press}
}

@misc{chie,
author = {Y. T. Chie},
title={Difference between Drawing from Photo Reference vs Real Life},
howpublished={\url{https://www.youtube.com/watch?v=XIhEiP0TvGE}}
}

@article{valueva2020application,
  title={Application of the residue number system to reduce hardware costs of the convolutional neural network implementation},
  author={Valueva, Maria V and Nagornov, NN and Lyakhov, Pavel Alekseevich and Valuev, Georgii V and Chervyakov, Nikolay I},
  journal={Mathematics and computers in simulation},
  volume={177},
  pages={232--243},
  year={2020},
  publisher={Elsevier}
}

@misc{sethi,
author = {R. J. Sethi},
title={Using computers to better understand art},
howpublished={\url{https://theconversation.com/using-computers-to-better-understand-art-56887}}
}

@inproceedings{bruno2013object,
  title={Object Recognition and Modeling Using SIFT Features},
  author={Bruno, Alessandro and Greco, Luca and La Cascia, Marco},
  booktitle={Advanced Concepts for Intelligent Vision Systems: 15th International Conference, ACIVS 2013, Pozna{\'n}, Poland, October 28-31, 2013. Proceedings 15},
  pages={250--261},
  year={2013},
  organization={Springer}
}

@article{krizhevsky2012imagenet,
  title={Imagenet classification with deep convolutional neural networks},
  author={Krizhevsky, Alex and Sutskever, Ilya and Hinton, Geoffrey E},
  journal={Advances in neural information processing systems},
  volume={25},
  year={2012}
}

@inproceedings{long2015fully,
  title={Fully convolutional networks for semantic segmentation},
  author={Long, Jonathan and Shelhamer, Evan and Darrell, Trevor},
  booktitle={Proceedings of the IEEE conference on computer vision and pattern recognition},
  pages={3431--3440},
  year={2015}
}

@inproceedings{redmon2016you,
  title={You only look once: Unified, real-time object detection},
  author={Redmon, Joseph and Divvala, Santosh and Girshick, Ross and Farhadi, Ali},
  booktitle={Proceedings of the IEEE conference on computer vision and pattern recognition},
  pages={779--788},
  year={2016}
}

@article{lowe2004distinctive,
  title={Distinctive image features from scale-invariant keypoints},
  author={Lowe, David G},
  journal={International journal of computer vision},
  volume={60},
  pages={91--110},
  year={2004},
  publisher={Springer}
}

@book{michelucci2019advanced,
  title={Advanced applied deep learning: convolutional neural networks and object detection},
  author={Michelucci, Umberto},
  year={2019},
  publisher={Springer}
}

@article{karami2017image,
  title={Image matching using SIFT, SURF, BRIEF and ORB: performance comparison for distorted images},
  author={Karami, Ebrahim and Prasad, Siva and Shehata, Mohamed},
  journal={arXiv preprint arXiv:1710.02726},
  year={2017}
}

@inproceedings{lin2015learning,
  title={Learning deep representations for ground-to-aerial geolocalization},
  author={Lin, Tsung-Yi and Cui, Yin and Belongie, Serge and Hays, James},
  booktitle={Proceedings of the IEEE conference on computer vision and pattern recognition},
  pages={5007--5015},
  year={2015}
}

@inproceedings{lowe1999object,
  title={Object recognition from local scale-invariant features},
  author={Lowe, David G},
  booktitle={Proceedings of the seventh IEEE international conference on computer vision},
  volume={2},
  pages={1150--1157},
  year={1999},
  organization={Ieee}
}

@article{bromley1993signature,
  title={Signature verification using a" siamese" time delay neural network},
  author={Bromley, Jane and Guyon, Isabelle and LeCun, Yann and S{\"a}ckinger, Eduard and Shah, Roopak},
  journal={Advances in neural information processing systems},
  volume={6},
  year={1993}
}

@inproceedings{melekhov2016siamese,
  title={Siamese network features for image matching},
  author={Melekhov, Iaroslav and Kannala, Juho and Rahtu, Esa},
  booktitle={2016 23rd international conference on pattern recognition (ICPR)},
  pages={378--383},
  year={2016},
  organization={IEEE}
}

@inproceedings{nguyen2016segmentation,
  author       = {Tam V. Nguyen and
                  Luoqi Liu and
                  Khang Nguyen},
  title        = {Exploiting generic multi-level convolutional neural networks for scene
                  understanding},
  booktitle    = {14th International Conference on Control, Automation, Robotics and
                  Vision, {ICARCV}},
  pages        = {1--6},
  publisher    = {{IEEE}}
}

@article{nguyen2019detection,
  author       = {Khanh{-}Duy Nguyen and
                  Khang Nguyen and
                  Duy{-}Dinh Le and
                  Duc Anh Duong and
                  Tam V. Nguyen},
  title        = {You always look again: Learning to detect the unseen objects},
  journal      = {Journal of Visual Communication and Image Representation},
  volume       = {60},
  pages        = {206--216},
  year         = {2019}
}

@article{nguyen2019saliency,
  author       = {Tam V. Nguyen and
                  Khanh{-}Duy Nguyen and
                  Thanh{-}Toan Do},
  title        = {Semantic Prior Analysis for Salient Object Detection},
  journal      = {{IEEE} Trans. Image Process.},
  volume       = {28},
  number       = {6},
  pages        = {3130--3141},
  year         = {2019}
}

\end{document}